\documentclass[sigconf,nonacm]{acmart}
\AtBeginDocument{%
  \providecommand\BibTeX{{%
    \normalfont B\kern-0.5em{\scshape i\kern-0.25em b}\kern-0.8em\TeX}}}

\acmConference{}
\acmYear{}
\copyrightyear{2022}
\setcopyright{none}

\acmBooktitle{} 
\acmPrice{}
\acmISBN{}

\begin{document}

\title{Neural apparent BRDF fields for multiview photometric stereo}


\author{Meghna Asthana}
\email{meghna.asthana@york.ac.uk}
\orcid{0000-0002-6101-8904}
\authornotemark[1]
\affiliation{%
  \institution{University of York}
  \city{York}
  \country{UK}
}

\author{William A.~P.~Smith}
\email{william.smith@york.ac.uk}
\orcid{0000-0002-6047-0413}
\affiliation{%
  \institution{University of York}
  \city{York}
  \country{UK}
}

\author{Patrik Huber}
\email{patrik.huber@york.ac.uk}
\orcid{0000-0002-1474-1040}
\affiliation{%
  \institution{University of York}
  \city{York}
  \country{UK}
}

\renewcommand{\shortauthors}{Asthana et al.}

\begin{abstract}
We propose to tackle the multiview photometric stereo problem using an extension of Neural Radiance Fields (NeRFs), conditioned on light source direction. The geometric part of our neural representation predicts surface normal direction, allowing us to reason about local surface reflectance. The appearance part of our neural representation is decomposed into a neural bidirectional reflectance function (BRDF), learnt as part of the fitting process, and a shadow prediction network (conditioned on light source direction) allowing us to model the apparent BRDF. This balance of learnt components with inductive biases based on physical image formation models allows us to extrapolate far from the light source and viewer directions observed during training. We demonstrate our approach on a multiview photometric stereo benchmark and show that competitive performance can be obtained with the neural density representation of a NeRF.

\end{abstract}




\begin{teaserfigure}
  \includegraphics[width=\textwidth]{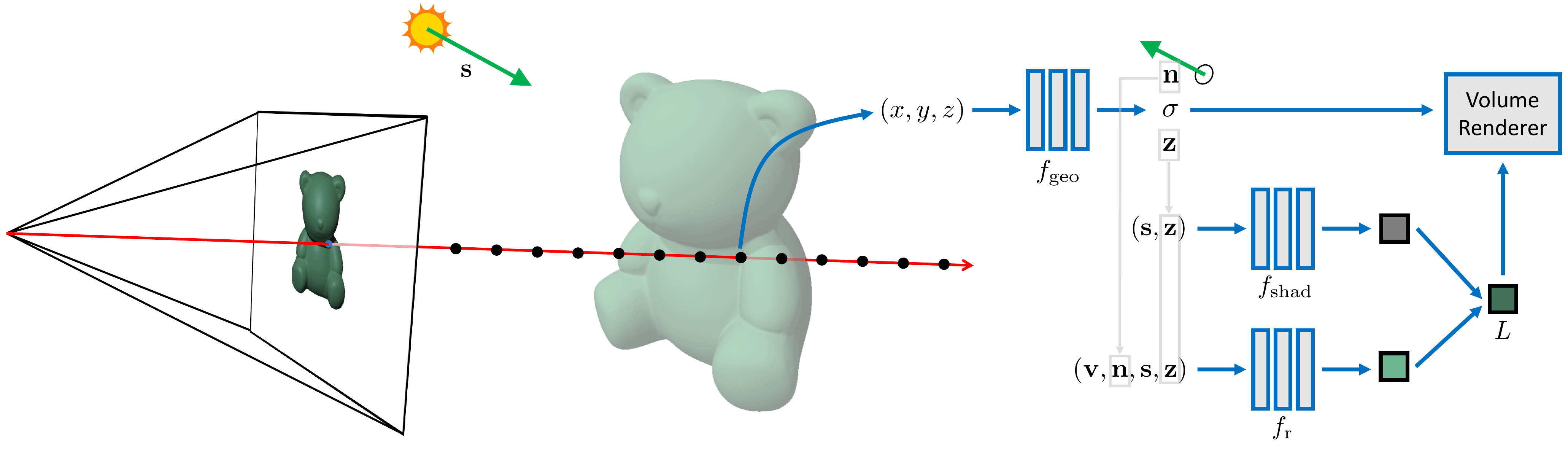}
  \caption{We replace the view-conditioned black box radiance predicted by a NeRF \cite{mildenhall2020nerf} with a physically-based image formation model. The geometry network predicts density and surface normal direction at each position in the volume. A neural BRDF and shadow network predict reflected scene radiance for the point which is then volume rendered using predicted density as for NeRF. The resulting model is relightable while also improving geometric information using multi-light observations.}
  \label{fig:overview}
\end{teaserfigure}

\maketitle

\section{Introduction}
\label{sec:intro}

Neural Radiance Fields (NeRFs) \cite{mildenhall2020nerf} use a coordinate-based multi-layer perceptron (MLP) to represent volumetric density and view-dependent radiance for a single scene. A NeRF can be rendered in a differentiable manner, meaning they can be trained from image data alone and they have proven to be an ideal representation for integrating information from multiple views. A trained NeRF is capable of state-of-the-art photorealistic novel view synthesis. Although much less explored, fitting such neural representations to images also provides a potential route to shape and material estimation.

In practice, a NeRF comprises two MLPs. The first is conditioned only on position and computes density. This can be thought of as capturing the geometry of the scene. The second is additionally conditioned on view direction. This network effectively learns the convolution of the bidirectional reflectance distribution function (BRDF) at each point in the scene with the lighting environment and non-local effects such as shadowing. In the specific case where lighting is provided by a single point source, it effectively learns a slice of the \emph{apparent BRDF} (i.e.~the BRDF combined with non-local effects) for each point in the scene in which lighting is fixed and viewpoint is allowed to vary over the full hemisphere.

In this paper, we develop this idea further. Specifically, we introduce explicit modelling of geometry and reflectance into a NeRF-based representation and use it to tackle the multiview photometric stereo problem. This problem provides both geometric- and illumination-calibrated images and we ask whether the NeRF representation is able to estimate accurate shape from such data. While a naive lighting dependency can be introduced by simply conditioning the second MLP also on light direction, this leads to very poor view synthesis extrapolation when the light source is far from the directions seen during training. Moreover, such an approach doesn't allow explicit modelling of geometry. 

Instead, as shown in Fig.~\ref{fig:overview}, we explicitly regress surface normal direction with the geometry network and decompose the second network into networks for BRDF (with suitable physical priors) and shadowing (i.e.~we learn the shadowing function for each point in the scene to avoid costly ray casting through the neural volume). We train the network using the same volume rendering and appearance losses as the original NeRF but use multi-illuminant data with known light source direction as in a photometric stereo setting. 

\begin{figure*}[!t]
  \centering
    \includegraphics[width=6in]{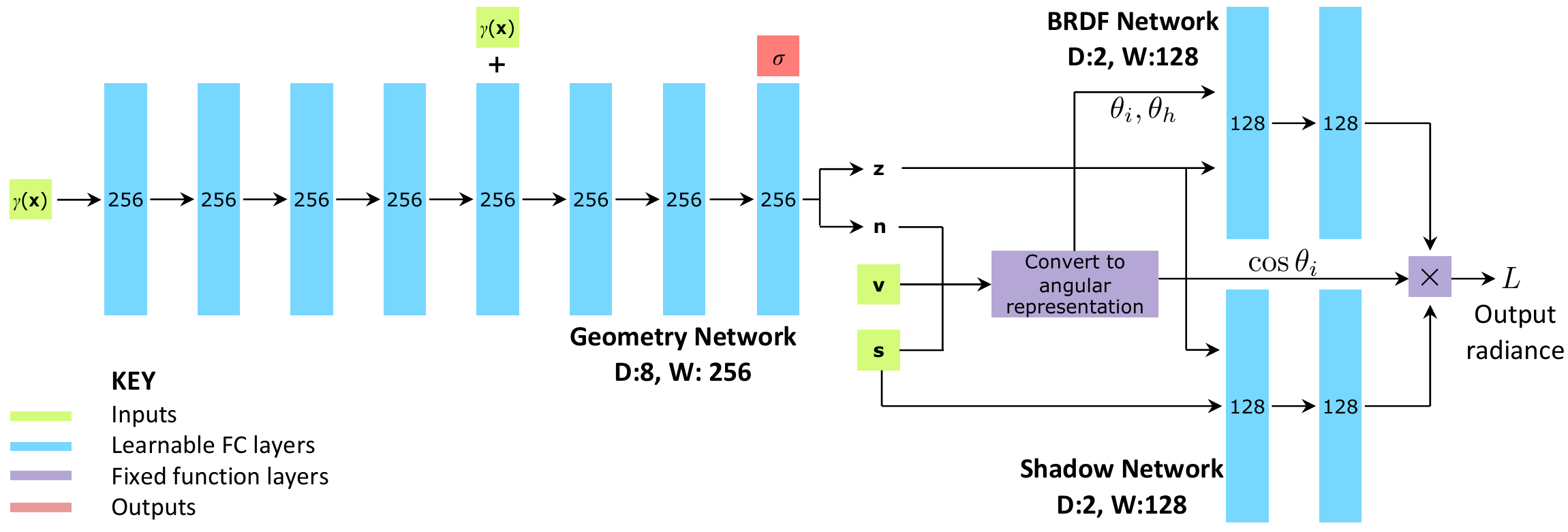}
  \caption{Our model is a combination of three learnable MLPs and fixed functions. The latent parameters $\mathbf{z}$ act as a bottleneck from which BRDF parameters and local shadowing function is inferred by the BRDF and shadow networks respectively.}
  \label{fig:architecture}
\end{figure*}

\section{Related work}

We begin by reviewing the Neural Radiance Fields method before describing extensions relevant to our work.

\paragraph{NeRF representation}
NeRF represents a scene using a neural volumetric representation. A continuous scene is represented as a 5D vector function with inputs 3D location $\mathbf{x} = (x, y, z)$ and 2D viewing direction $(\phi,\theta)$ (corresponding to a 3D Cartesian unit vector $\mathbf{d}$) which outputs an emitted colour $\mathbf{c}=(r,g,b)$ and volume density $\sigma$. The volume density function $\sigma$ is restricted to be a function only of location $\mathbf{x}$ in order to maintain multiview consistency. The network comprises a deep fully connected neural network without any convolutional layers (Multilayer Perceptron, MLP), $F_{\Theta}: (\mathbf{x,d})\mapsto(\mathbf{c,\sigma})$. The network outputs the radiance emitted in each direction at each point in space along with a density at each point which acts like a differential opacity controlling the amount of radiance which is accumulated when it passes through a particular point \cite{mildenhall2020nerf}. Previously, such MLP architectures have been used to represent natural textured materials which can be sampled over infinite domains \cite{henzler2020learning} or circumvent factors – like in Texture Fields – which limit the fidelity of highly textured surfaces \cite{oechsle2019texture}. 

\paragraph{Volume Rendering}
The network weights $\Theta$ are optimised such that reconstruction error to training images is minimised when the network output is passed to a differentiable volume renderer. Using the principles of classical volume rendering \cite{kajiya1984ray}, the colour of any ray passing through a scene can be rendered. The volume density $\sigma(\mathbf{x})$ can be interpreted as the differential probability of a ray terminating at an infinitesimal particle at location $\mathbf{x}$, then the expected colour $C(\mathbf{r})$ of camera ray $\mathbf{r}(t)=\mathbf{o}+t\mathbf{d}$ with near and far bounds $t_{n}$ and $t_{f}$ is:
\begin{equation}
    C(\mathbf{r})=\int_{t_n}^{t_f} T(t)\sigma(\mathbf{r}(t))\mathbf{c}(\mathbf{r}(t),\mathbf{d})dt,
\end{equation}
where
\begin{equation}
    T(t)=\exp\left(-\int_{t_n}^{t}\sigma(\mathbf{r}(s))ds\right).
\end{equation}
The function $T(t)$ represents the accumulated transmittance along the ray from $t_n$ to $t$ which is the probability of the ray travelling from $t_n$ to $t$ without hitting any other particle. This continuous integral is estimated using stratified sampling where $[t_n, t_f]$ is divided into $N$ evenly-spaced bins and values are randomly sampled from each bin as
\begin{equation}
    t_i\sim \mathcal{U}\left[t_n + \frac{i-1}{N}(t_f-t_n), t_n+\frac{i}{N}(t_f-t_n) \right].
\end{equation}
These samples are used to estimate $C(\mathbf{r}$) with the quadrature rule \cite{max1995optical} as
\begin{equation}
    \hat{C}(\mathbf{r})=\sum_{i=1}^{N}T_i(1-\mathrm{exp}(-\sigma_i\delta_i))\mathbf{c}_i,
\end{equation}
where
\begin{equation}
    T_i = \mathrm{exp}\left( -\sum_{j=1}^{i-1}\sigma_j\delta_j\right),
\end{equation}
where $\delta_i=t_{i+1}-t_i$ is the distance between adjacent samples and $\alpha_i=1-\mathrm{exp}(-\sigma_i\delta_i)$ represents alpha compositing.

\paragraph{Optimisation}
Volume rendering alone proves insufficient for achieving state-of-the-art rendering quality. The quality of complex scene representation in NeRF can be attributed to two factors - positional encoding and hierarchical volume sampling.

\begin{itemize}
    \item \underline{Positional Encoding:} It has been observed that the network $F_\Theta$ performs sub par when representing high-frequency variations in colour and geometry which can be attributed to deep networks being biased towards learning lower frequency functions \cite{rahaman2019spectral}. Therefore, operating on $xyz\theta\phi$ requires working with a composite function $F_\Theta=F_\Theta^{'}\circ\gamma$ ($\gamma$ is a mapping from $\mathbb{R}$ into higher dimensional space $\mathbb{R}^{2L}$) so as to enable the MLP to more easily approximate a higher frequency function \cite{mildenhall2020nerf}. $\gamma$ is defined as
    \begin{equation}
        \gamma(p)=(\sin{(2^0\pi p)}, \cos{(2^0\pi p)},\cdots ,\sin{(2^{L-1}\pi p)}, \cos{(2^{L-1}\pi p)}).
    \end{equation}
    
    \item \underline{Hierarchical Volume Sampling:} Evaluating the network at $N$ query points along every camera ray is computationally intensive and thus inefficient. For a more efficient evaluation, the rendering strategy \cite{levoy1990efficient,mildenhall2020nerf}, simultaneously optimises two networks - `coarse' and `fine'. A set of $N_c$ locations are used to first compute the `coarse' network. The output helps in informed sampling of relevant points for further detailed computation. The sole alternation is reconfiguration of alpha compositing colour from coarse network as
    \begin{equation}
        \hat{C}_c(\mathbf{r})=\sum_{i=1}^{N_c} w_i c_i, \; \; w_i=T_i(1-\mathrm{exp}(-\sigma_i\delta_i).
    \end{equation}
    \item \underline{Loss Function:} The loss function is defined as the total squared error between the rendered and true pixel colours for both coarse and fine renderings:
    \begin{equation}
        \mathcal{L}=\sum_{\mathbf{r}\in\mathcal{R}}\left[
        \begin{Vmatrix} 
            \hat{C}_c(\mathbf{r})-C(\mathbf{r})
        \end{Vmatrix}_2^2 +
        \begin{Vmatrix}
            \hat{C}_f(\mathbf{r})-C(\mathbf{r})
        \end{Vmatrix}_2^2\right]
    \end{equation}
    where $\mathcal{R}$ is the set of rays in each batch, and $\hat{C}(\mathbf{r})$, $\hat{C}_c(\mathbf{r})$, and $\hat{C}_f(\mathbf{r})$ are the ground truth, coarse volume predicted, and fine volume predicted RGB colours for ray $\mathbf{r}$ respectively.
\end{itemize}

\subsection{Neural 3D Representations}

Previous works in the field of neural rendering have learned shape priors from partial or noisy data. Local Implicit Grid Representations \cite{jiang2020local} trains an autoencoder to handle complex and diverse indoor scenes. Continuous Single Distance Function (DeepSDF) \cite{park2019deepsdf} provides trade-offs across fidelity, efficiency and compression capabilities enabling high quality shape representation, interpolation and completion. Local Deep Implicit Functions (LDIF) \cite{genova2020local} and Occupancy Networks \cite{mescheder2019occupancy} provide a computationally efficient function for high-resolution geometry representation for arbitrary topology. However, these representation functions are limited by a major factor – they all require access to ground truth of the 3D geometry which is typically obtained from a synthetic 3D shape.

A step up from these techniques have been in training reconstruction models from RGB images, however, these approaches have been restricted to voxel-~\cite{stutz2018learning,xie2019pix2vox,gadelha20173d} and mesh-based~\cite{kanazawa2018learning,liao2018deep,pan2019deep,wang2018pixel2mesh} representations which suffer from discretization or low resolution~\cite{niemeyer2020differentiable}. Recent works have provided a rather relaxed dependency on ground truth 3D geometry and essentially employing only 2D images to formulate differentiable rendering functions. Niemeyer et al.~\cite{niemeyer2020differentiable} propose to learn implicit shape and texture representations using depth gradients which are derived from implicit differentiation. This is obtained by calculating the surface intersection for each ray which is the input to a neural 3D texture field predicting a diffuse colour at a particular point. Sitzmann et al.~\cite{sitzmann2019scene} propose Scene Representation Networks (SRNs) which provides a structure-aware scene representation through a continuous function mapping world coordinates to features of a local scene – achieved through a differentiable ray-marching algorithm. These techniques have quite a potential to represent complicated and high-resolution geometry, they are still limited to extremely low geometric complexity leading to oversmoothed renderings.

\subsection{View Synthesis and Image-based Rendering}

Photorealistic novel views can be reconstructed when the data sampling is dense in nature by widely popular interpolation techniques, however, when the sampling is sparse in nature it becomes difficult to predict geometry and appearances from observed images. 
Mesh-based representations can be directly optimized using differentiable rasterizers and pathtracers. DIB-R \cite{chen2019learning} implements a differentiable renderer which views foreground rasterization as a weighted interpolation of local properties and background rasterization as a distance-based aggregation of global geometry – which, in turn, enables gradient calculation for all pixels in the image. Similarly, soft rasterizers \cite{liu2019soft} renders using an aggregation functions which collates probabilistic contributions of all mesh triangles with respect to the rendered pixels. Monte Carlo ray tracer \cite{li2018differentiable} through edge sampling provides a comprehensive solution to calculate scalar function derivates over a rendered image with respect to parameters like camera poses, lighting and scene geometry. However, these strategies require a template mesh with a fixed topology which is not possible for in-the-wild scenes.

Volumetric representations provide superior reconstruction for complex shapes and materials as they can be easily optimised through gradient-based approaches and are less prone to visually distracting artefacts. DeepView \cite{flynn2019deepview} synthesises multiplane images (MPI) from sparse camera viewpoints improving performance on object boundaries, light reflection and thin structures. Local light field fusion (LLFF) \cite{mildenhall2019local} proposed an algorithm by extending the traditional plenoptic sampling theory for view synthesis from an irregular grid of sample views via MPI representations. Neural Volumes \cite{lombardi2019neural} present an encoder-decoder framework which outputs a 3D volume representation from images and a differentiable ray-marching operation for end-to-end training. Though, these techniques have potential to high resolution novel view points, yet they are limited due to discrete sampling. 

\subsection{NeRF extensions}
The field of Neural Field Radiance has received significant attention due to their impressive results in capturing unbounded and bounded 360 degree scenes.  While NeRF works well on images of static subjects captured under controlled settings, it is incapable of modelling many ubiquitous, real-world phenomena in uncontrolled images, such as variable illumination or transient occluders \cite{martin2021nerf}. NeRF++ \cite{zhang2020nerf++} applies NeRF to 360 degree, large-scale, unbounded 3D scenes by applying shape radiance ambiguity and creating inverted sphere parameterization - separately modelling foreground and background. NeRF-W \cite{martin2021nerf} introduces a system for 3D reconstruction from in-the-wild photo collections. This was achieved by introducing Generative Latent Optimization (GLO) \cite{bojanowski2017optimizing} - making the approximation with respect to image-dependent radiance - and modelling observed colour as a probability distribution (isotropic normal distribution) over its value.  


A fundamental obstacle to making these methods practical is the extreme computational and memory requirements caused by the required volume integrations along the rendered rays during training and inference \cite{huang2020attention}. DeRF \cite{rebain2021derf} presents a solution to NeRF's computationally intensive nature using spatial decomposition where each portion of the scene is handled by smaller networks which can work together to render the complete scene. Furthermore, employing Voronoi spatial decomposition is efficient for GPU rendering and thus DeRF's inference is 3 times more efficient than NeRF. It also circumvents the issue of diminishing returns in neural rendering. Neural Sparse Voxel Fields (NSVF) \cite{liu2020neural} provided faster (20x) and high quality free-viewpoint rendering using voxel-bounded implicit fields. These model local properties of each cell by learning the underlying voxel structure - using a differentiable ray-marching operation - and skipping ones which do not provide any scene content. AutoInt \cite{autoint} presents an efficient learning framework using implicit neural networks. When applied to neural volume rendering, it provides improved computational efficiency. 


Closely related to this paper have been several previous efforts to incorporate physically-based image formation models into NeRF-like architectures. These differ in their modelling assumptions, input data and target applications.

Neural Reflectance Decomposition (NeRD) \cite{boss2021nerd} presents a model which decomposes a scene into its shape, reflectance and illumination. NeRD achieves this by introducing physically-based rendering with light independent reflectance parameters instead of direct view-dependent colour. The BRDF is estimated using additional network which creates severe bottleneck i.e.~a two-dimensional latent space encoding every possible BRDFs in the scene. As the underlying shape and BRDF are preemptively assumed to be same for all the views, the model converges to a consistent state for input images with varied illumination. Neural Reflectance and Visibility Fields (NeRV) \cite{srinivasan2021nerv} describes a model which can render a 3D representation for novel viewpoints under arbitrary lighting conditions from a set of input images with unconstrained known lighting conditions. It takes 3D location as input to learn scene properties like volume density, surface normal, material parameters, distance to the first surface intersection and visibility of external environment in any direction. Altogether, they not only provide novel views but also indirect illumination effect which makes the model's performance superior to prior approaches.

Neural Radiance Factorization (NeRFactor) \cite{zhang2021nerfactor} addresses the issue of recovering the shape and spatial-varying reflectance of an object from multi-view images of an object illuminated by unknown lighting condition which enables the rendering of novel views of the object under arbitrary environment lighting and object's material properties. The major technical contribution include, a method for factorizing images of an object under an unknown lighting condition into shape, reflectance and illumination i.e.~creating free-view point relighting (with shadows) and material editing, a strategy to distill NeRF-estimated volume density into surface geometry to use for initialisation and novel data-driven BRDF prior learned from training a latent code model on real measured BRDFs.

The input for the NeRFactor model are multi-view images of an object lit by an unknown illumination condition. After optimization, NeRFactor outputs, $\mathbf{x}$, a 3D location on the object surface, $\mathbf{n}$ surface normal, $\mathbf{v(\omega_i)}$ light visibility in any direction, $\mathbf{a}$ albedo and $\mathbf{z_{BRDF}}$ reflectance. By explicitly modeling light visibility, NeRFactor is able to separate shadows from albedo and synthesize realistic soft or hard shadows under arbitrary lighting conditions. NeRFactor is able to recover convincing 3D models for free-viewpoint relighting in this challenging and under constrained capture setup.

The Neural Reflectance and Visibility Fields (NeRV) \cite{srinivasan2021nerv} provides a method to render 3D representation of any scene from a novel viewpoint under arbitrary lighting conditions. The NeRV architecture takes 3D locations as input and provides multiple scene properties like volume density, surface normal, material parameters, distance to the first surface intersection in any direction, and visibility of the external environment in any direction. These parameters are crucial for the illumination performance of the model.
NeRV implementation has replaced NeRF's radiance MLP with two MLPs: a "shape" MLP which outputs volume density $\sigma$ and a "reflectance" MLP which outputs BRDF parameters (3D diffused albedo $\mathbf{a}$ and one-dimensional roughness $\gamma$) for any input 3D point. Furthermore, they have introcued a "visibility" MLP for estimating general lighting scenarios (which is computationally challenging) which emits an approximation of the environment lighting visibility at any input location along any direction. 

In this paper, we propose a novel combination of directly regressed surface normals, neural BRDF, neural shadows and the application to multiview photometric stereo data. We believe we are the first to present results on real multiview photometric stereo images (as opposed to synthetic data) and to show convincing normal map reconstruction from such data, devoid of the high frequency noise typical in previous work that uses normals differentiated from density.



\begin{figure*}[!t]
    \centering
    \begin{tabular}{c@{\hspace{0.05cm}}c@{\hspace{0.05cm}}c@{\hspace{0.05cm}}c}
    {\scriptsize Rendering} & {\scriptsize Closest training image} & {\scriptsize Surface Normals} & {\scriptsize Shadows} \\
        \includegraphics[width=4cm,clip=true,trim=170px 180px 150px 101px]{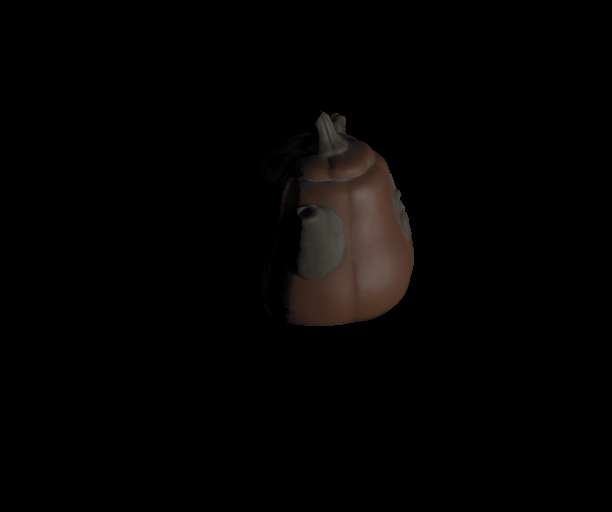} & 
        \includegraphics[width=4cm,clip=true,trim=170px 180px 150px 101px]{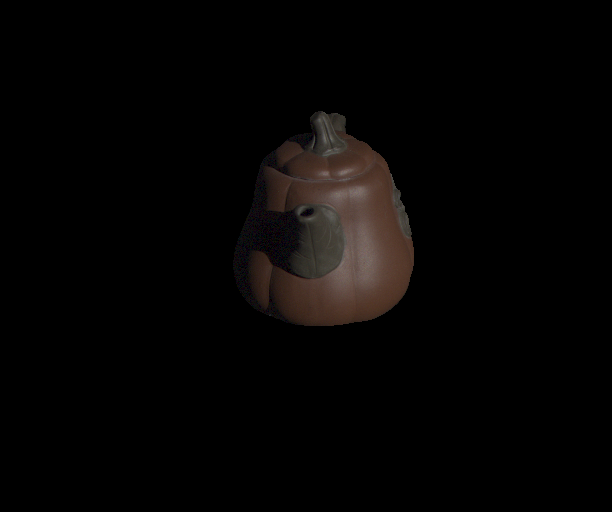} & 
        \includegraphics[width=4cm,clip=true,trim=170px 180px 150px 101px]{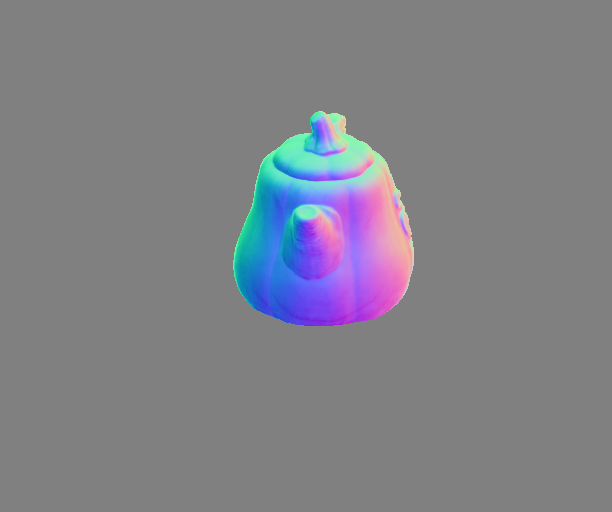} &
        \includegraphics[width=4cm,clip=true,trim=170px 180px 150px 101px]{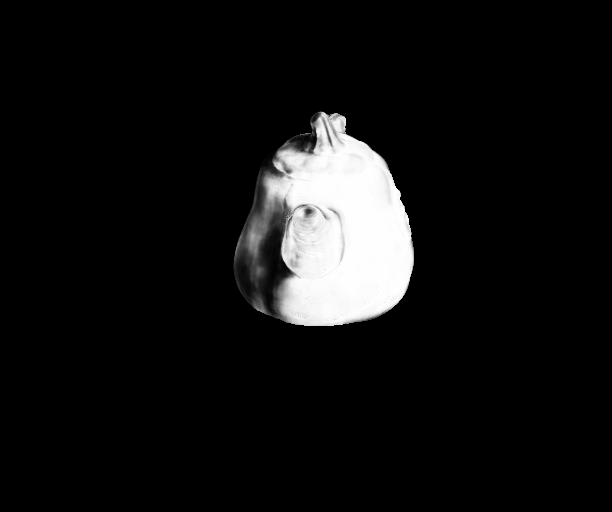} \vspace{-0.05cm} \\ 
        \includegraphics[width=4cm,clip=true,trim=160px 168px 160px 25px]{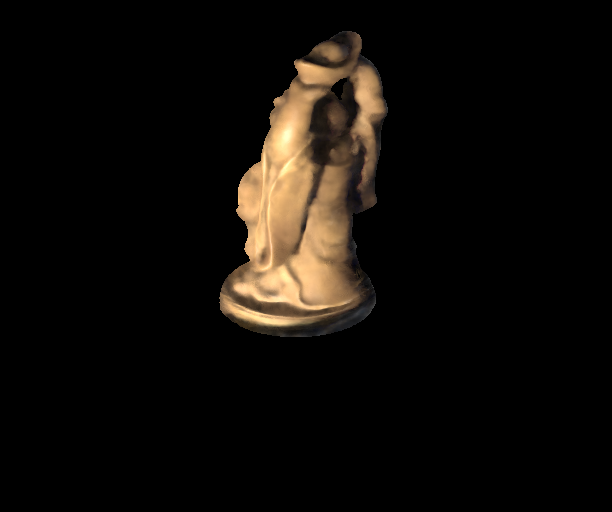} & 
        \includegraphics[width=4cm,clip=true,trim=160px 168px 160px 25px]{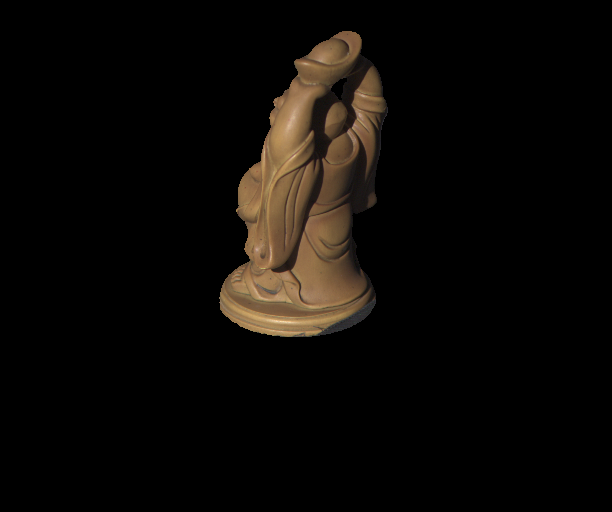} & 
        \includegraphics[width=4cm,clip=true,trim=160px 168px 160px 25px]{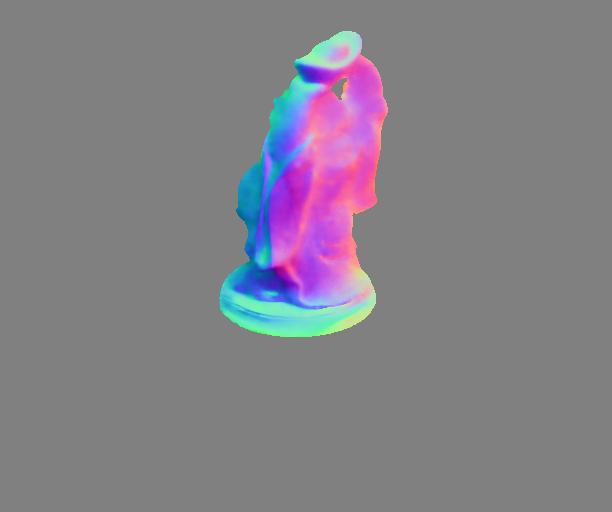} &
        \includegraphics[width=4cm,clip=true,trim=160px 168px 160px 25px]{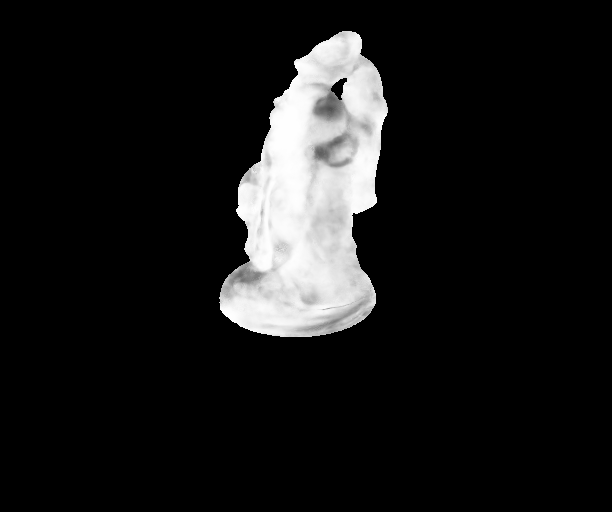} \vspace{-0.05cm} \\
        \includegraphics[width=4cm,clip=true,trim=160px 178px 160px 68px]{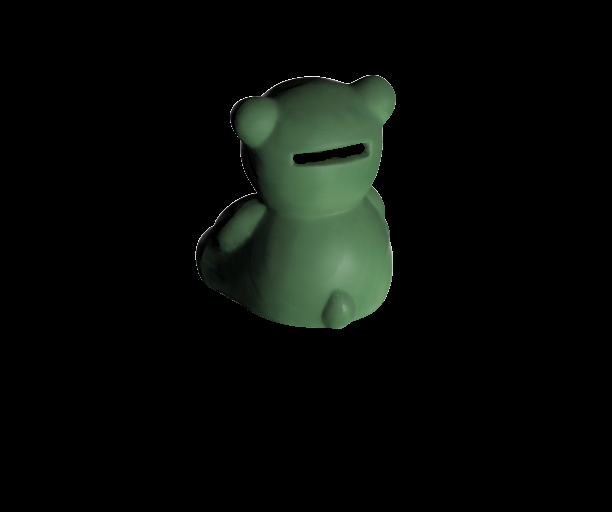} & 
        \includegraphics[width=4cm,clip=true,trim=160px 178px 160px 68px]{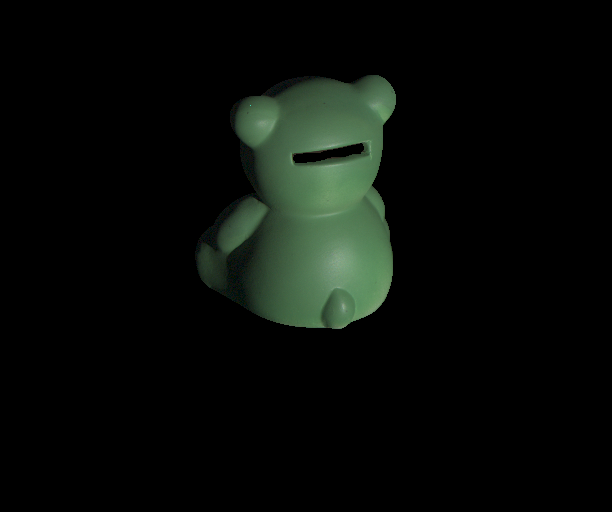} & 
        \includegraphics[width=4cm,clip=true,trim=160px 178px 160px 68px]{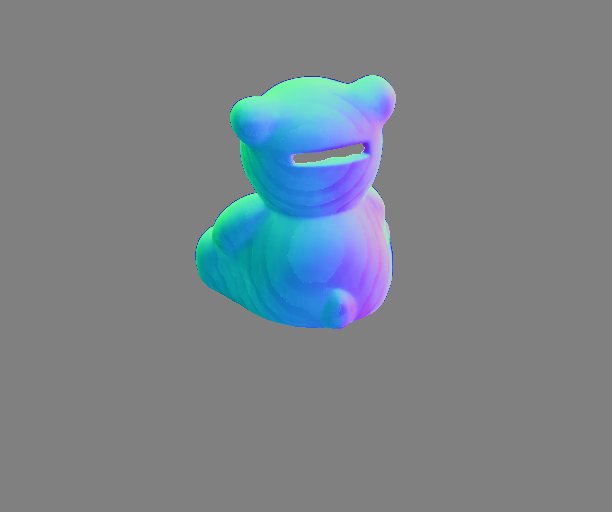} &
        \includegraphics[width=4cm,clip=true,trim=160px 178px 160px 68px]{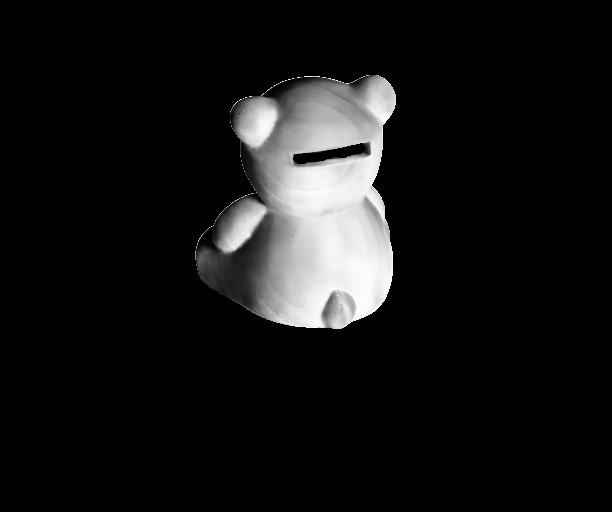} \vspace{-0.05cm} \\
        \includegraphics[width=4cm,clip=true,trim=160px 151px 160px 68px]{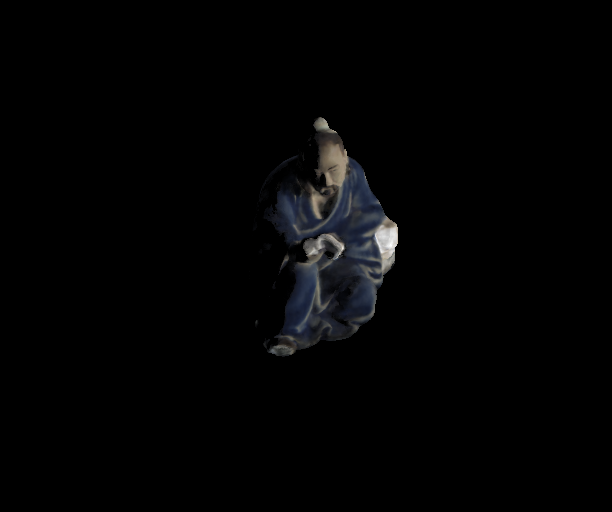} & 
        \includegraphics[width=4cm,clip=true,trim=160px 151px 160px 68px]{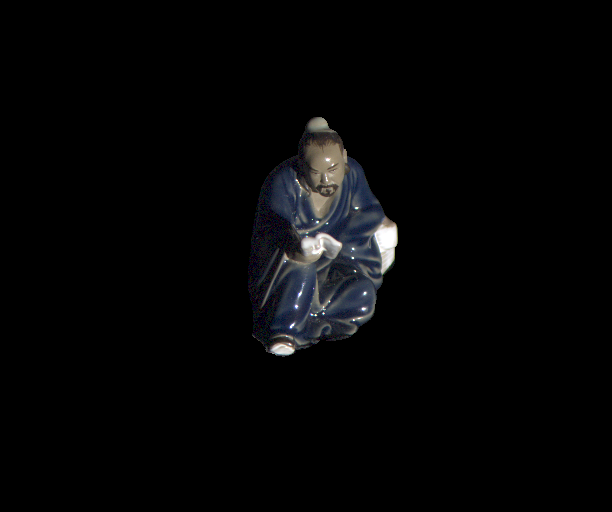} & 
        \includegraphics[width=4cm,clip=true,trim=160px 151px 160px 68px]{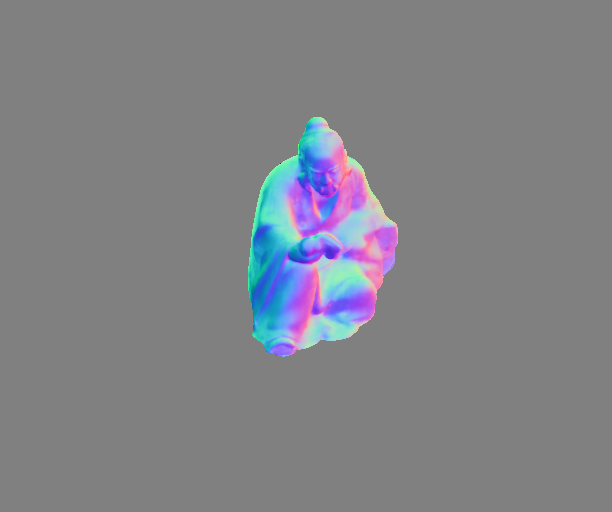} &
        \includegraphics[width=4cm,clip=true,trim=160px 151px 160px 68px]{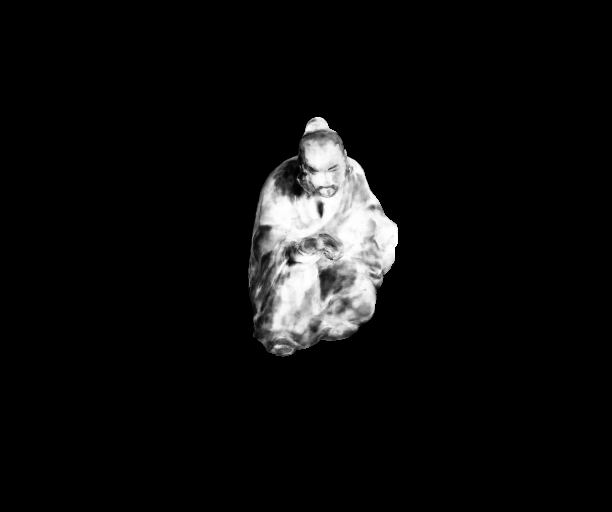} \vspace{-0.05cm} \\
        \includegraphics[width=4cm,clip=true,trim=160px 178px 160px 120px]{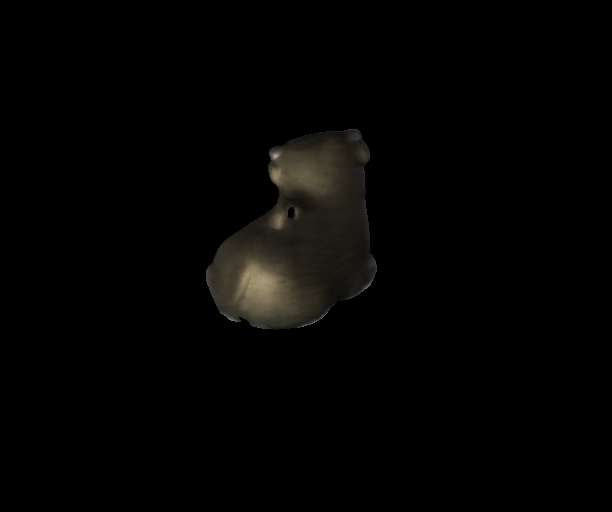} & 
        \includegraphics[width=4cm,clip=true,trim=160px 178px 160px 120px]{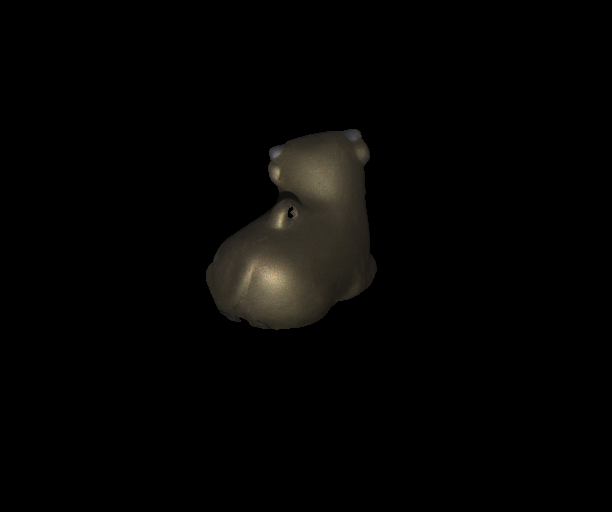} & 
        \includegraphics[width=4cm,clip=true,trim=160px 178px 160px 120px]{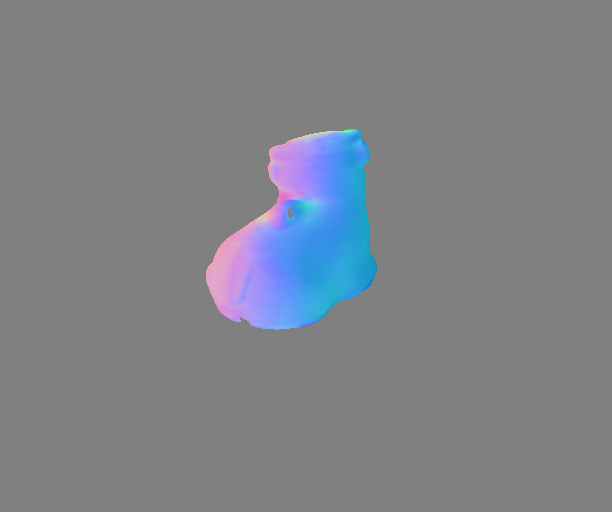} &
        \includegraphics[width=4cm,clip=true,trim=160px 178px 160px 120px]{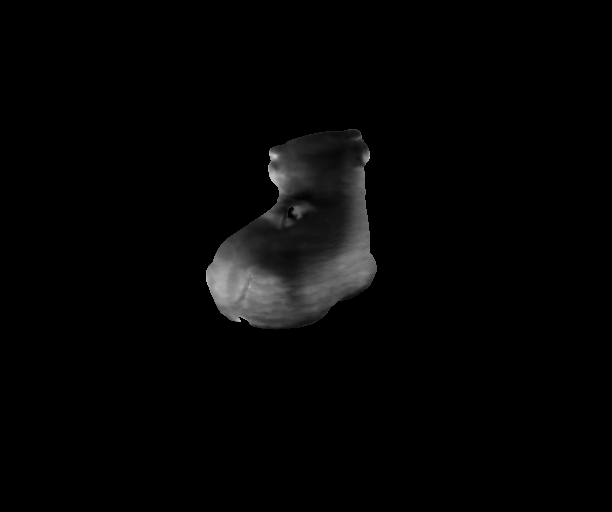} \vspace{-0.05cm} \\
    \end{tabular}
    \caption{Qualitative results for Diligent-MV~\cite{li2020multi} using neural BRDF and shadow networks. From left to right: rendering from the trained model under novel lighting and viewpoint, closest training image, estimated surface normal and shadow maps.}
    \label{fig:diligent_qualitative_brdf}
\end{figure*}

\section{Neural Apparent BRDF Fields}

The original NeRF model can be written as two MLPs: 
\begin{enumerate}
\item $f^{\Theta_{\text{geo}}}_{\text{geo}}:\mathbf{x}\mapsto (\mathbf{z},\sigma)$ with learnable parameters $\Theta_{\text{geo}}$ which maps a 3D location $\mathbf{x}=(x,y,z)$ to a latent code $\mathbf{z}\in\mathbb{R}^d$ and density $\sigma\in\mathbb{R}_{\geq 0}$.
\item $f^{\Theta_{\text{col}}}_{\text{col}}:(\mathbf{z},\mathbf{v})\mapsto \mathbf{c}$ with learnable parameters $\Theta_{\text{col}}$ which maps the viewing direction $\mathbf{v}\in\mathbb{R}^3$, $\|\mathbf{v}\|=1$, and latent code to a colour $\mathbf{c}=(r,g,b)$. 
\end{enumerate}
A naive relightable extension to NeRF simply conditions the colour network on a point source light direction, $\mathbf{s}\in\mathbb{R}^3$, $\|\mathbf{s}\|=1$, i.e.~replace the second network with: $f^{\Theta_{\text{col}}}_{\text{col}}:(\mathbf{z},\mathbf{v},\mathbf{s})\mapsto \mathbf{c}$. Note that the latent code $\mathbf{z}$ must store an encoding of the apparent BRDF at a given point in the scene. Appearance under arbitrary illumination environments can be obtained by integrating the output of this network over incident illumination directions. To train such a network we require training images with both viewpoint variation and variation in illumination direction provided by a point light source of known direction. 

Such a naive extension neglects several underlying physical properties of surface reflectance which strongly limits its ability to accurately interpolate and extrapolate to new lighting directions. We therefore propose a series of modifications to create a NeRF architecture that follows more closely a physical image formation model.

\subsection{Geometry network}

Surface reflectance is determined by the local surface orientation where the light strikes the surface. To be able to reason about this explicitly, first, we extend the initial geometry network to regress not only density but also surface normal direction: $f^{\Theta_{\text{geo}}}_{\text{geo}}:\mathbf{x}\mapsto (\mathbf{z},\mathbf{n},\sigma)$, where $\mathbf{n}\in\mathbb{R}^3$, $\|\mathbf{n}\|=1$, is the unit length surface normal (in practice we regress an unconstrained 3D vector and normalise to unit length in-network). An alternative is to derive the normal from the gradient of the density field as done in \cite{srinivasan2021nerv}. This explicitly ties the two geometric representations together. However, we found that allowing the network to directly regress surface normal direction enables reconstruction of higher frequency details in the normal map and reduces noise artefacts in the normal maps caused by using second order derivatives. This is better suited to the task of multiview photometric stereo where high quality per-view normal maps may be a useful output in their own right.

\subsection{Neural BRDF}

Second, we use a network to explicitly model the BRDF: $f^{\Theta_{\text{r}}}_{\text{r}}:(\mathbf{n},\mathbf{s},\mathbf{v},\mathbf{z})\mapsto \rho$, where $\rho\in\mathbb{R}^3$ represents the BRDF value for each of the RGB channels. The latent code $\mathbf{z}$ is used to encode information about the BRDF at the scene point. The learnable parameters of the network, $\Theta_{\text{r}}$, are optimised such that the network can represent the range of BRDFs observed in the scene, as parameterised by $\mathbf{z}$. A BRDF is not an arbitrary function - it must obey some physical constraints. We satisfy two of these: 1.~positivity: satisfied by using an activation function that can only output positive values, 2.~reciprocity: we parameterise the incident and outgoing directions using an angular representation that is itself reciprocal (i.e.~unchanged by switching $\mathbf{s}$ and $\mathbf{v}$). Satisfying the third physical constraint, conservation of energy, we leave as future work due to the difficulty of imposing integral constraints on neural networks.

As a special case, this network can be a fixed function representing a known parametric BRDF. In this case, we use a learnable network to map the latent code to the parameters of the BRDF: $f_{\text{param}}:\mathbf{z}\mapsto \mathbf{p}$, where $\mathbf{p}\in\mathbb{R}^P$ contains the specific parameters for the chosen BRDF model. As a specific example, the Lambertian model has $P=3$ parameters for the RGB diffuse albedo. Then the BRDF becomes a non-learnable function $f_{\text{r}}:(\mathbf{n},\mathbf{s},\mathbf{v},\mathbf{p})\mapsto \rho$. We explore this variant in our experimental results.

\subsection{Shadow prediction}

Modelling reflectance using a BRDF means we could only describe local reflectance effects. We equip our model with additional power to allow it to describe the effect of shadows. These are the most significant non-local effect encountered in our data. The combination of BRDF and shadowing means we model a partially decomposed apparent BRDF. Our neural BRDF model will learn some non-local effects such as interreflection while the gross changes caused by cast shadows can be explained by the shadow network.

Of course, exact cast shadows can be computed from a NeRF by ray casting, but this is expensive. So, we follow the idea in NeRV~\cite{srinivasan2021nerv} and instead replace ray casting by a learnt approximation. Specifically, we train an MLP that predicts a scalar soft shadow value from light source direction and the NeRF latent code: $f^{\Theta_{\text{shad}}}_{\text{shad}}:(\mathbf{s},\mathbf{z})\mapsto [0,1]$. For fixed latent code, this MLP then represents the light source visibility function.

\subsection{Camera behaviour and nonlinear appearance loss}

Surface reflectance can have high dynamic range (for example, specularities on glossy objects can be orders of magnitude brighter than diffuse reflections). For this reason, similar to NeRF in the dark \cite{mildenhall2021rawnerf} we let our model learn high dynamic range radiance in a linear space. We then apply nonlinear tone-mapping (in our case, a simple gamma correction) before computing the error to tone-mapped training images. This significantly outperforms computing errors in linear space where the bright regions completely dominate the error. Operating in a nonlinear tone-mapped space ensures the reconstruction better approximates minimisation of a perceptual error.

\subsection{Rendering}

In order to compute the reflected radiance at scene point $\mathbf{x}$ of a single white point light source from direction $\mathbf{s}$ of unit intensity towards the viewer in direction $\mathbf{v}$, our pointwise rendering equation is:
\begin{equation}
    L^{\Theta}(\mathbf{x},\mathbf{s},\mathbf{v}) = f^{\Theta_{\text{r}}}_{\text{r}}(\mathbf{n}(\mathbf{x}),\mathbf{s},\mathbf{v},\mathbf{z}(\mathbf{x}))f^{\Theta_{\text{shad}}}_{\text{shad}}(\mathbf{s},\mathbf{z}(\mathbf{x}))\max(0,\cos\theta_i),
\end{equation}
where $(\mathbf{n}(\mathbf{x}),\mathbf{z}(\mathbf{x}))=f^{\Theta_{\text{geo}}}_{\text{geo}}(\mathbf{x})$, $\theta_i=\arccos(\mathbf{n}(\mathbf{x})\cdot \mathbf{s})$ and $\Theta=(\Theta_{\text{geo}},\Theta_{\text{r}},\Theta_{\text{shad}})$ is all the learnable parameters in the model. We use this reflected radiance quantity in place of the view-conditioned radiance in the original NeRF model and volume render using the estimated density $\sigma$ in the same way, i.e.
\begin{equation}
    \hat{C}(\mathbf{r})=\sum_{i=1}^{N}T_i(1-\mathrm{exp}(-\sigma_i\delta_i))L^{\Theta}(\mathbf{x}_i,\mathbf{s},\mathbf{v}).
\end{equation}

\subsection{Mask supervision}

For individual objects with known foreground masks, we find performance is substantially improved by supervising the network using the mask. This takes two forms. First, we can penalise any density along rays that do not pass through the object mask. We use the image masks to divide the set of rays $\mathcal{R}$ into two disjoint subsets for object foreground $\mathcal{F}$ and background $\mathcal{B}$. We then introduce a silhouette loss that encourages zero density outside the visual hull of the object:
\begin{equation}
    \mathcal{L}_{\text{sil}} = \sum_{\mathbf{r}\in\mathbb{B}} \sum_{i=1}^N \sigma_i.
\end{equation}
This is equivalent to the shape-from-silhouette cue and serves to directly constrain the geometry of the object. Second, we mask the appearance loss to only penalise appearance errors inside the object:
$
    \mathcal{L}_{\text{app}} = \sum_{\mathbf{r}\in\mathbb{F}} \| \hat{C}(\mathbf{r})^{1/2.2} - C(\mathbf{r}) \|^2,
$
where $C(\mathbf{r})$ are the ground truth values and we use a fixed gamma of $2.2$.

\begin{figure*}[!t]
    \centering
    \begin{tabular}{c@{\hspace{0.05cm}}c@{\hspace{0.05cm}}c@{\hspace{0.05cm}}c@{\hspace{-0.05cm}}c}
    {\scriptsize Rendering} & {\scriptsize Closest training}\vspace{-0.1cm} & {\scriptsize Surface Normals} & {\scriptsize Albedo} & {\scriptsize Shadows} \\
     & {\scriptsize image} &  &  &  \\
        \includegraphics[width=3.2cm,clip=true,trim=160px 178px 160px 68px]{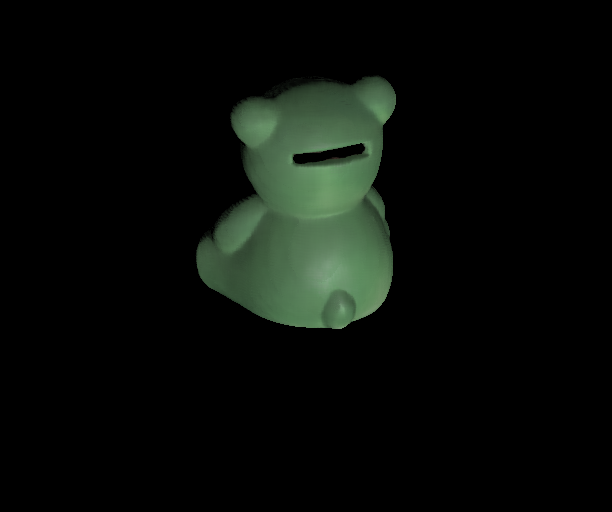} & 
        \includegraphics[width=3.2cm,clip=true,trim=160px 178px 160px 68px]{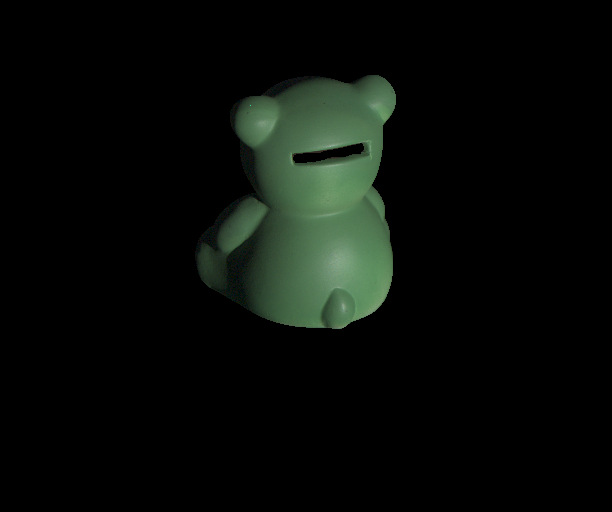} & 
        \includegraphics[width=3.2cm,clip=true,trim=160px 178px 160px 68px]{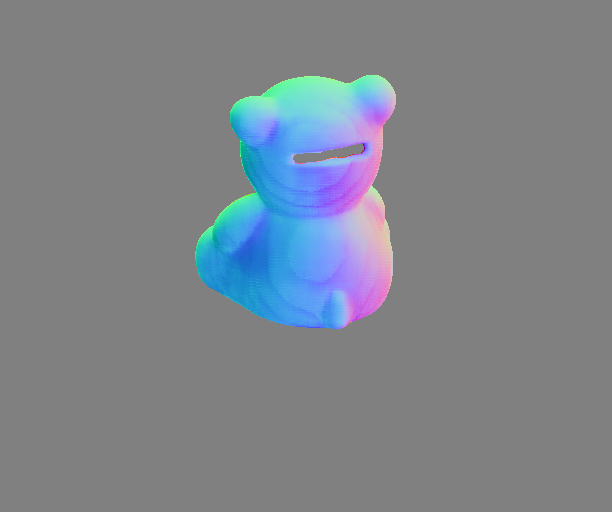} &
        \includegraphics[width=3.2cm,clip=true,trim=160px 178px 160px 68px]{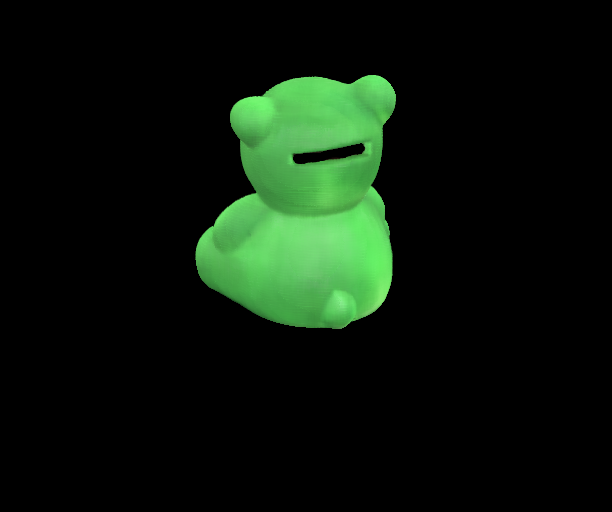} \vspace{-0.05cm} &
        \includegraphics[width=3.2cm,clip=true,trim=160px 178px 160px 68px]{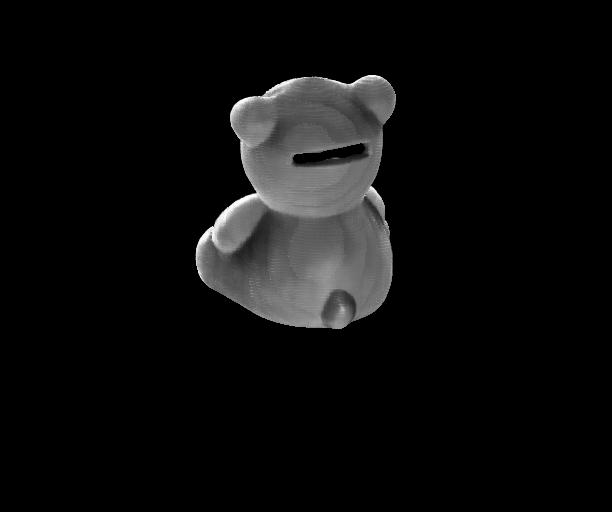} \vspace{-0.05cm} \\
        
        \includegraphics[width=3.2cm,clip=true,trim=160px 151px 160px 68px]{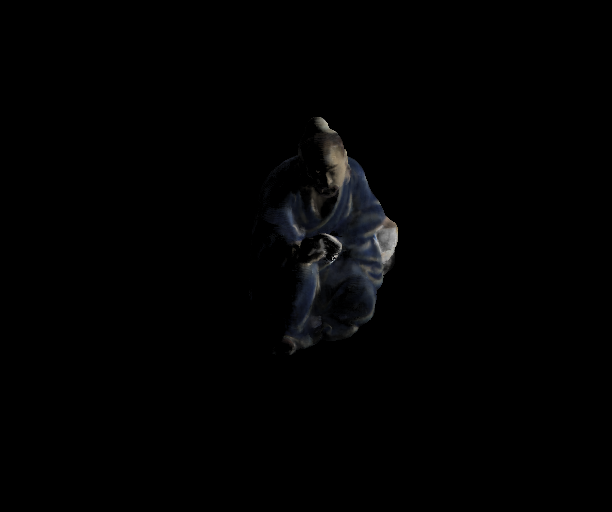} &
        \includegraphics[width=3.2cm,clip=true,trim=160px 151px 160px 68px]{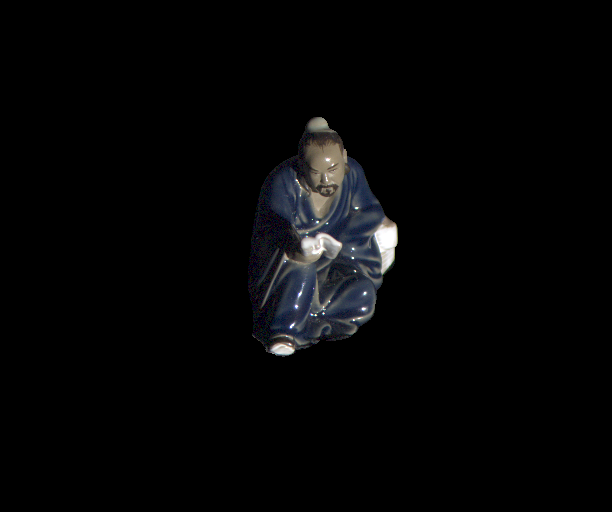} &
        \includegraphics[width=3.2cm,clip=true,trim=160px 151px 160px 68px]{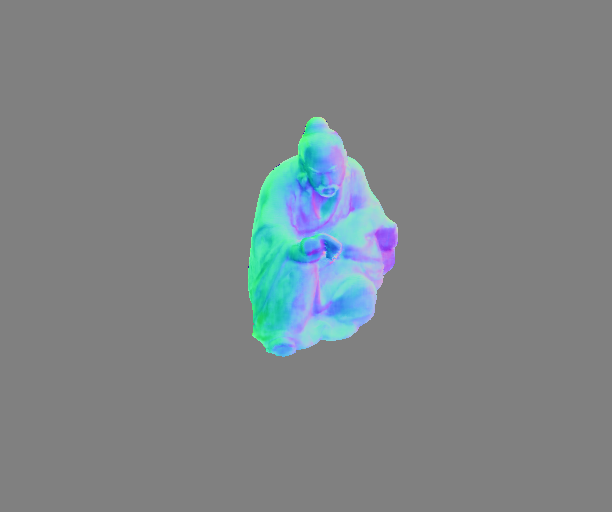} &
        \includegraphics[width=3.2cm,clip=true,trim=160px 151px 160px 68px]{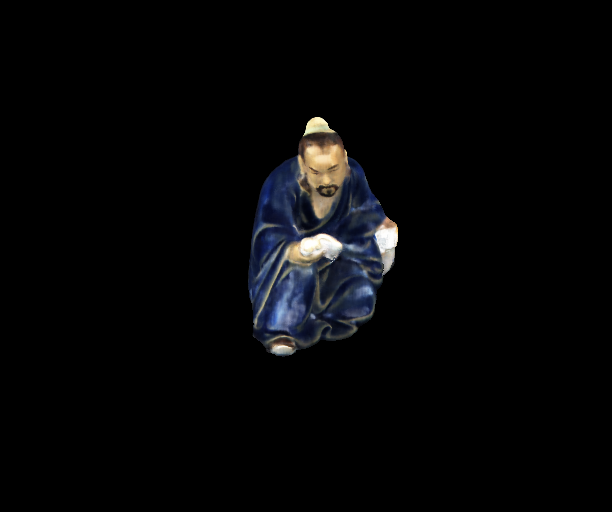} \vspace{-0.05cm} &
        \includegraphics[width=3.2cm,clip=true,trim=160px 151px 160px 68px]{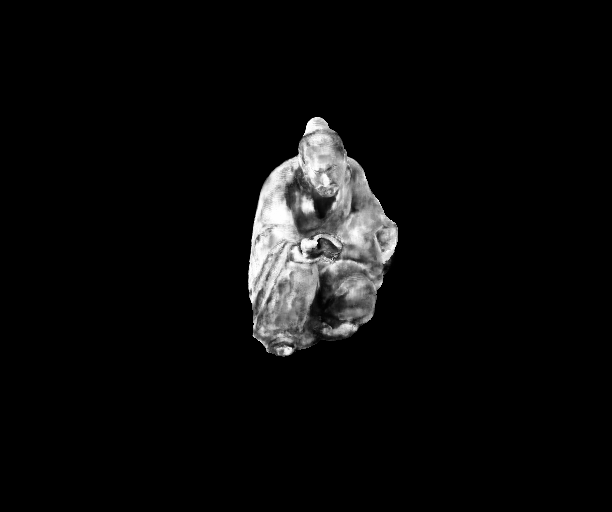} \vspace{-0.05cm} \\
        
        \includegraphics[width=3.2cm,clip=true,trim=160px 178px 160px 120px]{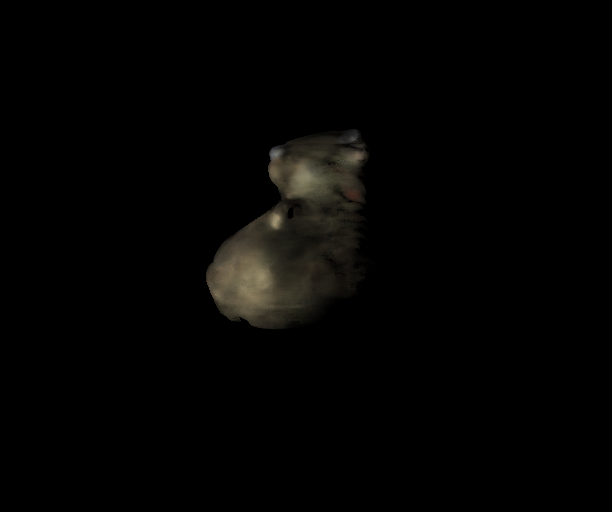} & 
        \includegraphics[width=3.2cm,clip=true,trim=160px 178px 160px 120px]{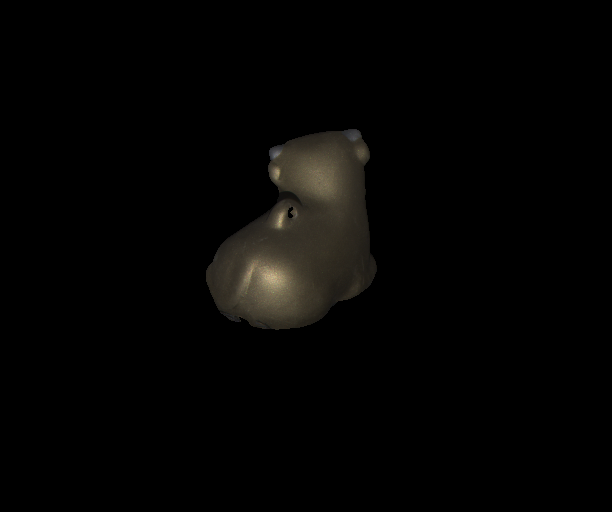} & 
        \includegraphics[width=3.2cm,clip=true,trim=160px 178px 160px 120px]{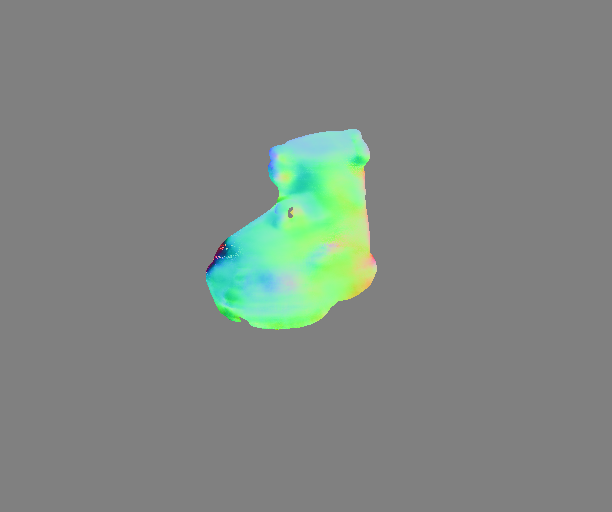} &
        \includegraphics[width=3.2cm,clip=true,trim=160px 178px 160px 120px]{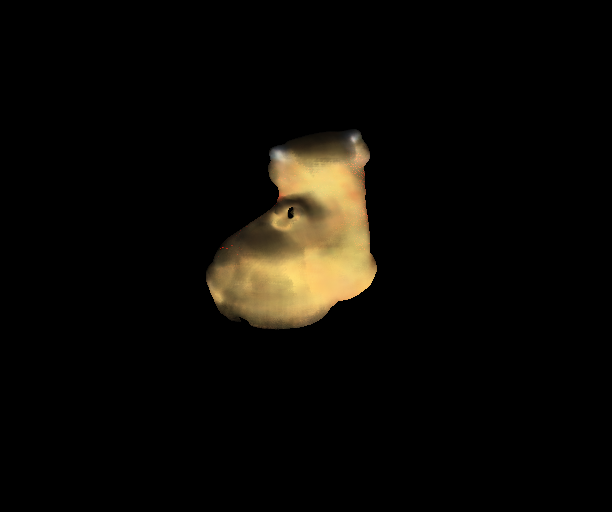} \vspace{-0.05cm} &
        \includegraphics[width=3.2cm,clip=true,trim=160px 178px 160px 120px]{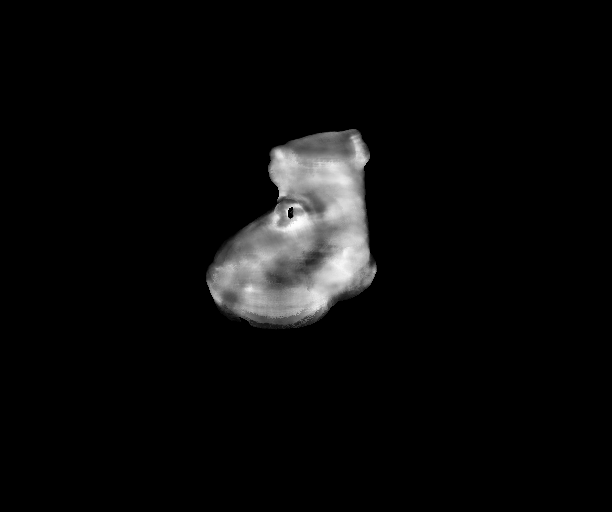} \vspace{-0.05cm} \\
    \end{tabular}
    \caption{Qualitative results for Diligent-MV~\cite{li2020multi} using Lambertian and shadow networks. From left to right: rendering from the trained model under novel lighting and viewpoint, closest training image, estimated surface normal, diffuse albedo and shadow maps.}
    \label{fig:diligent_qualitative_lambertian}
\end{figure*}

\begin{figure*}
    \centering
    \begin{tabular}{c@{\hspace{0.05cm}}c@{\hspace{0.05cm}}c@{\hspace{0.05cm}}c@{\hspace{0.05cm}}c@{\hspace{0.05cm}}c}
        \includegraphics[width=2.5cm,clip=true,trim=175px 156px 175px 101px]{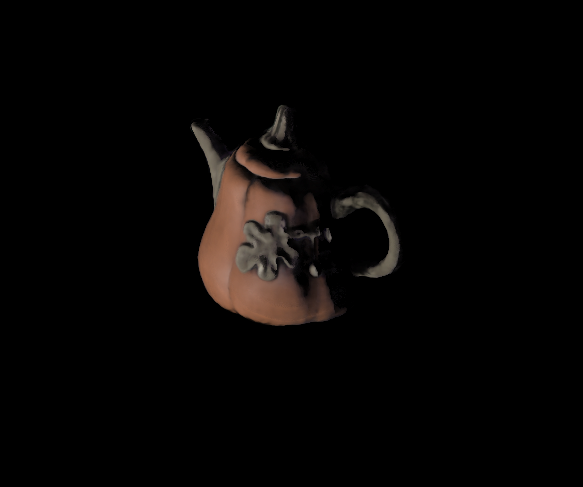} & 
        \includegraphics[width=2.5cm,clip=true,trim=175px 156px 175px 101px]{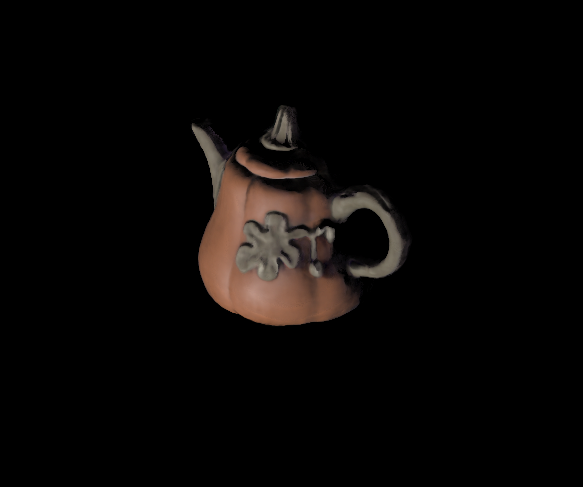} & 
        \includegraphics[width=2.5cm,clip=true,trim=175px 156px 175px 101px]{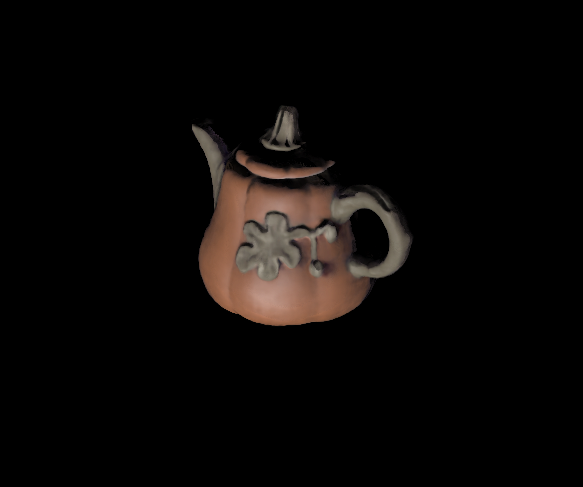} & 
        \includegraphics[width=2.5cm,clip=true,trim=175px 156px 175px 101px]{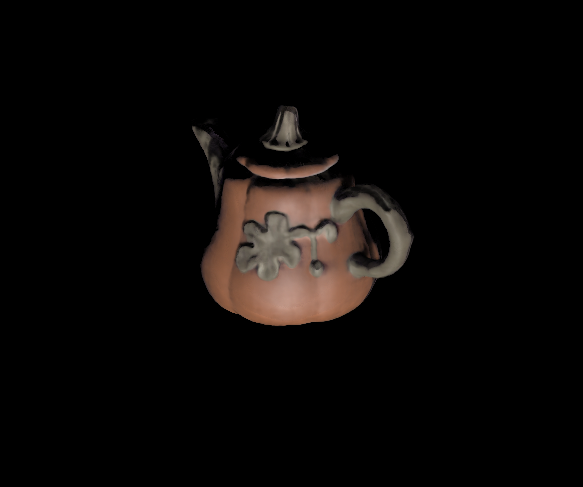} & 
        \includegraphics[width=2.5cm,clip=true,trim=175px 156px 175px 101px]{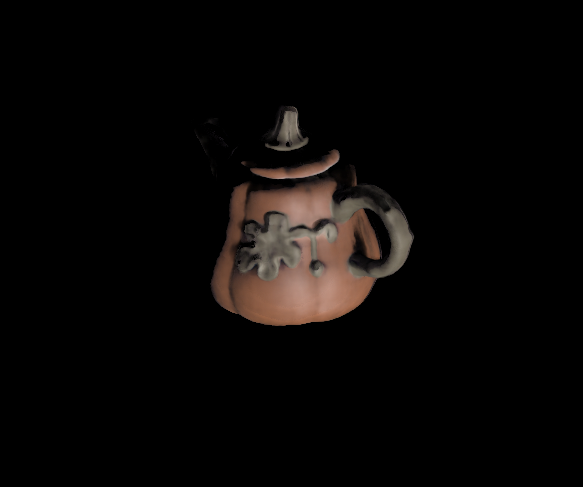} & 
        \includegraphics[width=2.5cm,clip=true,trim=175px 156px 175px 101px]{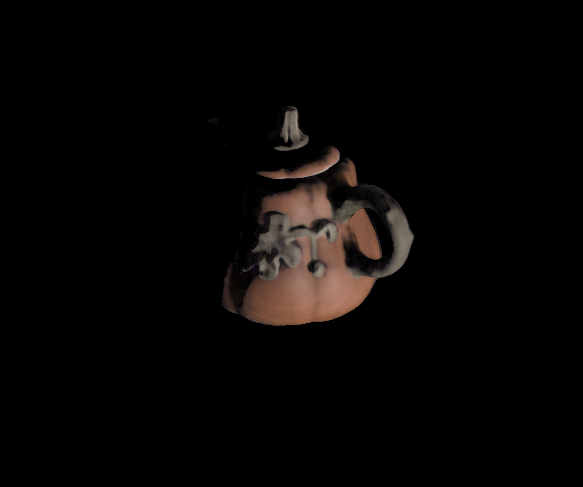} \\
        \includegraphics[width=2.5cm,clip=true,trim=175px 156px 175px 101px]{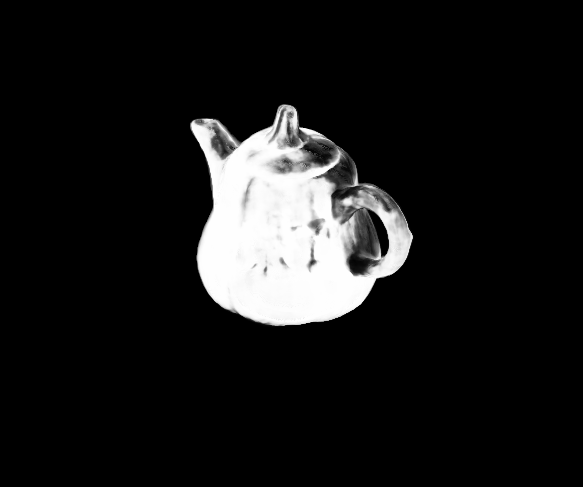} & 
        \includegraphics[width=2.5cm,clip=true,trim=175px 156px 175px 101px]{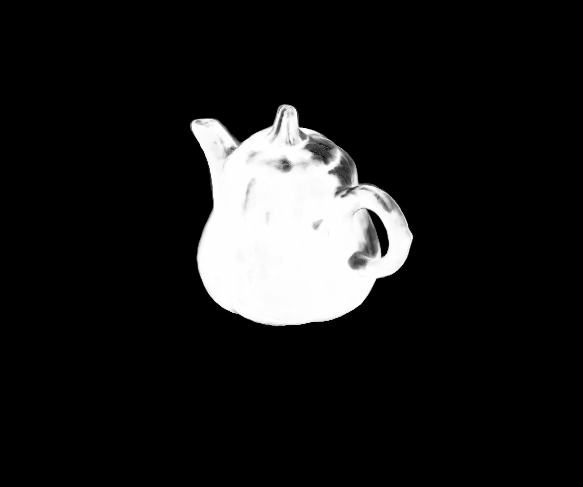} & 
        \includegraphics[width=2.5cm,clip=true,trim=175px 156px 175px 101px]{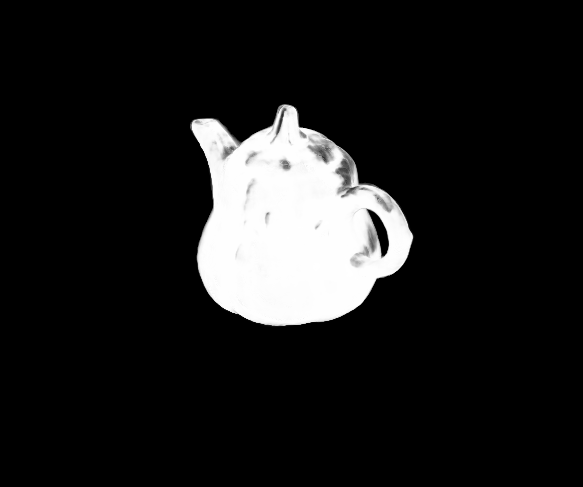} & 
        \includegraphics[width=2.5cm,clip=true,trim=175px 156px 175px 101px]{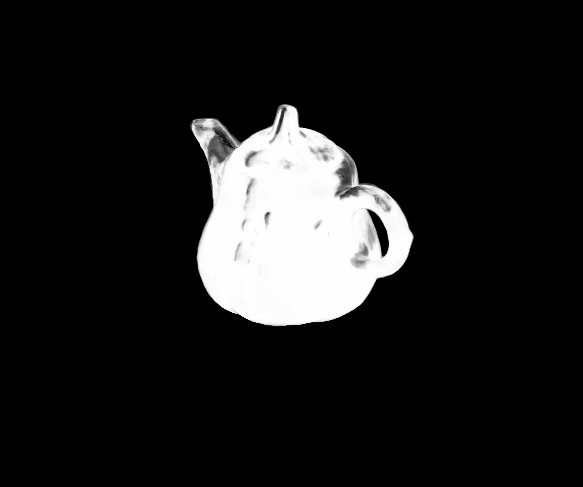} & 
        \includegraphics[width=2.5cm,clip=true,trim=175px 156px 175px 101px]{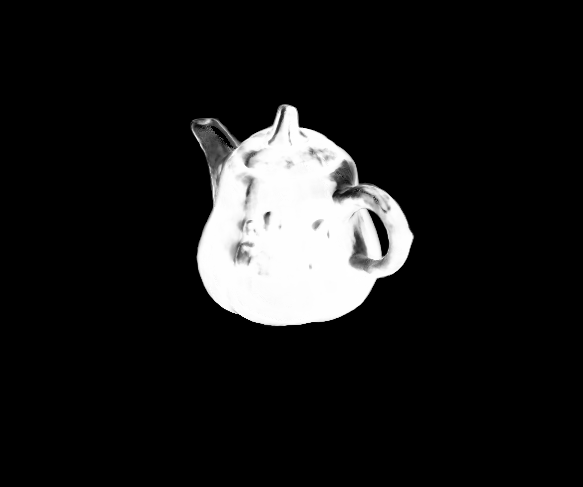} & 
        \includegraphics[width=2.5cm,clip=true,trim=175px 156px 175px 101px]{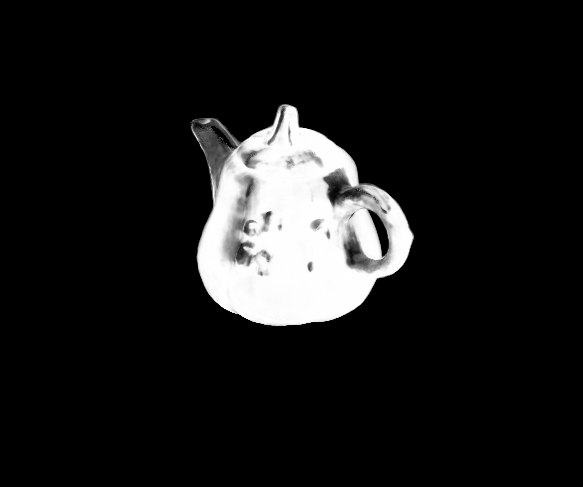} \\
    \end{tabular}
    \caption{Relighting result (top) and the corresponding rendering of the shadow prediction only (bottom).}
    \label{fig:relighting}
\end{figure*}

\begin{figure*}
    \centering
    \resizebox{\textwidth}{!}{
    \begin{tabular}{cccccccc}
    \includegraphics[width=1cm,trim=120px 52px 56px 40px,clip=true]{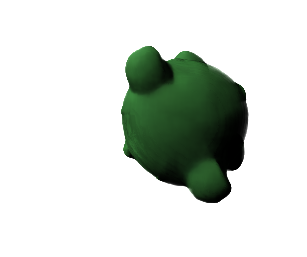} &
    \includegraphics[width=1cm,trim=120px 52px 56px 40px,clip=true]{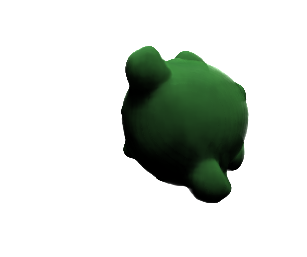} &
    \includegraphics[width=1cm,trim=120px 52px 56px 40px,clip=true]{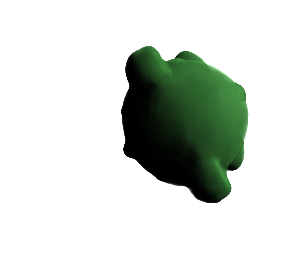} &
    \includegraphics[width=1cm,trim=120px 52px 56px 40px,clip=true]{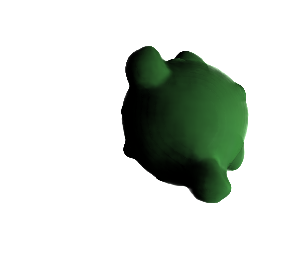} &
    \includegraphics[width=1cm,trim=120px 52px 56px 40px,clip=true]{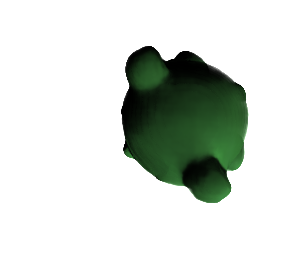} &
    \includegraphics[width=1cm,trim=120px 52px 56px 40px,clip=true]{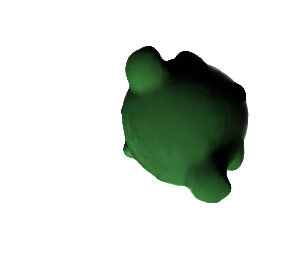} &
    \includegraphics[width=1cm,trim=120px 52px 56px 40px,clip=true]{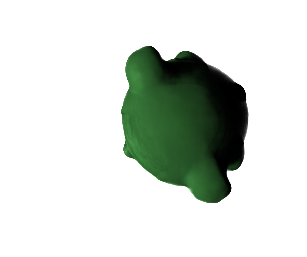} &
    \includegraphics[width=1cm,trim=120px 52px 56px 40px,clip=true]{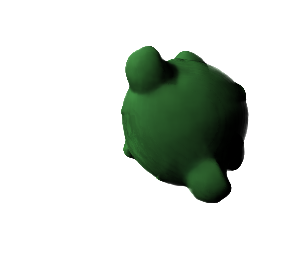} \\
    \end{tabular}
    }
    \caption{Relighting under extreme pose and illumination extrapolation. We show a top down view of the bear object with 360\textdegree{} light rotation around the object.}
    \label{fig:360relight}
\end{figure*}

\begin{table*}[!t]
\scriptsize
    \centering
    \resizebox{\textwidth}{!}{
    \begin{tabular}{@{}llllllllllllllllllllll@{}}
    \toprule
    & \multicolumn{20}{c}{\textbf{Pose}} & \\
    & 01 & 02 & 03 & 04 & 05 & 06 & 07 & 08 & 09 & 10 & 11 & 12 & 13 & 14 & 15 & 16 & 17 & 18 & 19 & 20 & Mean\\
    \midrule
    \textbf{Pot}     & 13.4 & 13.1 & 13.2 & 13.3 & 13.0 & 12.4 & 11.6 & 11.1 & 11.1 & 11.1 & 11.3 & 11.8 & 12.2 & 12.6 & 12.7 & 12.7 & 12.7 & 12.2 & 12.3 & 12.4 & 12.3 \\
    \textbf{Bear}    & 22.3 & 22.2 & 22.6 & 22.5 & 21.8 & 20.5 & 19.4 & 19.0 & 19.5 & 20.6 & 22.1 & 22.9 & 23.7 & 24.0 & 24.0 & 23.7 & 23.1 & 22.2 & 21.7 & 21.3 & 22.0 \\
    \textbf{Buddha}  & 22.9 & 23.7 & 23.4 & 24.4 & 24.5 & 26.3 & 29.2 & 27.7 & 25.9 & 21.4 & 18.5 & 19.7 & 22.7 & 25.8 & 26.5 & 26.4 & 25.7 & 23.2 & 22.3 & 24.2 & 24.2 \\
    \textbf{Reading} & 22.4 & 18.3 & 18.5 & 17.5 & 17.4 & 17.8 & 17.3 & 17.2 & 17.0 & 17.0 & 16.9 & 16.8 & 16.0 & 16.0 & 15.8 & 15.5 & 15.4 & 15.4 & 15.4 & 16.0 & 17.0 \\
    \textbf{Cow}     & 47.6 & 51.1 & 59.5 & 65.4 & 65.8 & 64.3 & 63.7 & 62.2 & 57.7 & 50.6 & 49.4 & 56.1 & 65.3 & 69.9 & 66.4 & 61.5 & 65.6 & 69.3 & 64.5 & 53.7 & 60.5 \\
    \bottomrule
    \end{tabular}
    }
    \vspace{0.0em} 
    \caption{Surface normal estimation errors. For each object in each pose, we render normal maps from our neural model and compute mean angular error over the foreground region. In the last column we show results averaged over poses.}
    \label{tab:angular_errors}
\end{table*}

\section{Implementation}

In Figure \ref{fig:architecture} we show the overall architecture of our model. For the geometry network, we follow NeRF and use an MLP with 8 layers, 256 hidden units per layer and ReLU activation. The dimensionality of the latent vector $\mathbf{z}$ is 256. Both the BRDF and shadow networks are MLPs with 2 layers, 128 hidden units per layer and ReLU activation. For the surface normal $\mathbf{n}$, we regress an unconstrained 3D vector then rescale to unit length. For our angular representation of BRDF geometry, we use a simple two angle representation: the incident angle $\theta_i=\arccos \mathbf{n}\cdot \mathbf{s}$ and the half angle $\theta_h=\arccos \mathbf{n}\cdot \mathbf{h}$ where $\mathbf{h}=(\mathbf{s}+\mathbf{v})/\|\mathbf{s}+\mathbf{v}\|$ is the half way vector. For the variant with fixed (Lambertian) BRDF, we replace the BRDF network with a single linear layer to map $\mathbf{z}$ to 3 values and use sigmoid activation to compute RGB diffuse albedo.

We follow the original NeRF and apply positional encoding to input position $\mathbf{x}$. Since we use view direction $\mathbf{v}$ for the computation of angles for BRDF calculation, we do not apply positional encoding to this. We also follow the original NeRF optimisation in using both a coarse and fine network combined with stratified sampling.

We work entirely in the normalised NeRF coordinate system. This rescales camera poses such that the object lies approximately inside the unit cube with cameras lying approximately around the equator. We convert light source directions to this coordinate system meaning that the normal maps we estimate are in NeRF world coordinates. Where needed, we convert these to camera coordinates via the rotational component of the camera extrinsics.

\section{Experiments}

\paragraph{Dataset}
We use the DiLiGenT-MV dataset \cite{li2020multi}. This is a multiview photometric stereo dataset containing images of $5$ objects with complex BRDFs taken from 21 different calibrated viewpoints. The images are originally captured in 16-bit HDR which we tonemap to LDR using a fixed gamma of 2.2. Images are of resolution $612\times 512$ and illuminated from 96 different light source directions. However, the lights are positioned in 2 rings around the camera and only cover a small area of the incident hemisphere. The maximum angle any light source direction makes with the view vector is $42.9^{\circ}$. This provides a good test case for extrapolating appearance when the light source direction is further from the viewer. The dataset is provided with ground truth masks which we use in our silhouette loss and masked appearance loss.

\paragraph{View synthesis and relighting} Our trained model can be used to synthesise novel views with arbitrary (potentially unseen) viewpoint and lighting direction. In Figure~\ref{fig:diligent_qualitative_brdf}, column 1 we show rendering results of our model. For comparison, we show the closest image in the training set in the column 2. Note that we are able to produce more extreme lighting directions that predict realistic shadows and position of specularities. In columns 3 and 4 we show the normal and shadow maps obtained by volume rendering the normal and shadow predictions. Figure \ref{fig:diligent_qualitative_lambertian} shows qualitative results for the variant in which we used a fixed (Lambertian) BRDF. In this case, the model predicts only diffuse albedo parameters with no learnable BRDF. We retain the learnable shadow network. While these results provide plausible albedo estimates, they are not able to synthesise non-Lambertian effects such as specular reflections. In Figure~\ref{fig:relighting} we show the effect of systematically varying light source direction as the light rotates around the front of the object. We show shadow maps in the second column. Note the prediction of the cast shadow by the handle onto the teapot. We emphasise that this dependency is learnt and not ray cast. Finally in Figure \ref{fig:360relight} we show an extreme extrapolation example. The training views form a ring around the equator of the object. We synthesise a top down view 90 degrees from the training view directions. Similarly, the synthesised 360 rotating light is 90 degrees from the virtual view direction, much greater than observed during training.

\paragraph{Shape estimation} In Table~\ref{tab:angular_errors} we show quantitative shape estimation errors in terms of angular error between estimated and ground truth surface normal maps. The normal maps can also be inspected visually in Figures \ref{fig:diligent_qualitative_brdf} and \ref{fig:diligent_qualitative_lambertian}. While the numerical errors are larger than state-of-the-art methods, particularly for the more challenging materials, our qualitative results show that we are able to estimate detailed and smooth shape using our model.

\section{Conclusion}

In this paper we have presented the first attempt to use the neural volumetric density representation of NeRF combined with neural estimates of intrinsic geometric and material surface properties for the task of multiview photometric stereo. We have shown that the approach is feasible and allows view synthesis extrapolation far from the viewpoints and light source directions seen in the training data. While the metric accuracy of the shape estimates does not yet match that of hand engineered multiview photometric stereo solutions, we believe there is potential to do so. We believe that the current accuracy is held back by the soft density representation. For objects comprising hard surfaces, a representation that directly represents surfaces is likely to provide better results. For example, the combination of a neural implicit surface with volume rendering~\cite{wang2021neus} shows great promise and could be easily combined with our image formation model and rendering process. Imposing additional constraints on the neural BRDF, either by explicitly enforcing conservation of energy or by learning a shared model across multiple objects or even a BRDF database is also likely to help constrain the problem and reduce ambiguities. Finally, the angular input to the neural BRDF is a topic for further exploration. We used a simple incident angle/half angle representation which can model basic non-Lambertian phenomena. Alternatives such as the Rusinkiewicz angles~\cite{rusinkiewicz1998new} might offer improved BRDF modelling.

\bibliographystyle{splncs04}
\bibliography{refs}

\end{document}